\pdfoutput=1
\documentclass[11pt]{article}

\usepackage[]{ACL2023}

\usepackage{times}
\usepackage{latexsym}
\usepackage{multirow}
\usepackage{graphicx}

\usepackage[T1]{fontenc}
\usepackage[utf8]{inputenc}

\usepackage{microtype}

\usepackage{inconsolata}
\usepackage{amsmath}
\usepackage{todonotes}
\newcommand*\iftodonotes{\if@todonotes@disabled\expandafter\@secondoftwo\else\expandafter\@firstoftwo\fi}  % defines \iftodonotes{<true>}{<false>}, thanks to \makeatother

\usepackage{changes}
\usepackage[nameinlink]{cleveref}

\title{Exploring the Limitations of Detecting Machine-Generated Text}

\author{
  Jad Doughman\textsuperscript{1}, 
  Osama Mohammed Afzal\textsuperscript{1}, 
  Hawau Olamide Toyin\textsuperscript{1}, \\
  {\bf Shady Shehata\textsuperscript{1}, 
  Preslav Nakov\textsuperscript{1}, 
  Zeerak Talat\textsuperscript{2}} \\
  \textsuperscript{1}Mohamed bin Zayed University of Artificial Intelligence \textsuperscript{2}University of Edinburgh\\
  \texttt{\{jad.doughman, preslav.nakov\}@mbzuai.ac.ae, z@zeerak.org}
}

\begin{document}
\maketitle
\begin{abstract}
Recent improvements in the quality of the generations by large language models have spurred research into identifying machine-generated text. 
Such work often presents high-performing detectors. 
However, humans and machines can produce text in 
different 
styles and 
domains, yet the the performance impact of such on
machine generated text detection systems remains unclear.
In this paper, we 
audit the classification performance for detecting machine-generated text by evaluating on texts with varying writing styles.
We find that classifiers are highly sensitive to stylistic changes and differences in text complexity, and in some cases degrade entirely to random classifiers. 
We further find that detection systems are particularly susceptible to misclassify easy-to-read texts while they have high performance for complex texts, leading to concerns about the reliability of detection systems. 
We recommend that future work attends to stylistic factors and reading difficulty levels of human-written and machine-generated text.

%\looseness=-1
% Our findings suggest that the performance of detectors for machine generated text are highly dependent on domain and text complexity.   
% We call for the research community to perform more careful data analysis.
% to afford more accurate reporting.\looseness=-1

%Large Language Models (LLMs) have evolved to be able to generate coherent passages with varying writing styles, all while being grounded by external data sources for improved factuality. These improvements have made the distinction between machine-generated and human written passages ever so difficult as neither writing style, coherence, nor factuality can act as distinguishing features. Human's writing styles are also infinitely varied, including differences in sentence length, vocabulary, tone, and structure. This poses a question on the separability of the classes as well as the generalizability of existing detection classifiers. To this extent, this paper aims to investigate the limitations of detecting machine generated text by evaluating classifiers on sentences with varying writing styles and computing performance drops across each subgroup. Our findings illustrate that classifiers are highly sensitive to the distribution of verbs, adverbs, and average sentence length with an F-score reduction to near zero levels across both machine-generated and human written classes. Additionally, impact of reading ease + lexical diversity/sophistication + domain shift + punctuation marks and whitespaces.
\end{abstract}

\section{Introduction}
% TODO: ADD CITATIONS

Recent developments for % advancements in 
large language models (LLMs) have enabled the generation of text that mimics human writing in %both 
coherence and style, which can be used for benign (e.g., drafting an e-mail) %presenting an e-mail draft) 
or for nefarious (e.g., %to generate 
generating misinformation at scale) purposes. 
%This has allowed bad actors to leverage these capabilities to rapidly produce and spread misinformation, facilitate plagiarism (generate text similar to existing works), and amplify biases present in training data. 
To mitigate the risks of machine-generated text (MGT), research has devoted efforts 
%the Natural Language Processing (NLP) community has devoted efforts 
to building MGT detectors~\cite[e.g.,][]{wang-etal-2024-m4,Koike_Kaneko_Okazaki_2024, info14100522}.%for MGT~\cite[e.g.,][]{wang-etal-2024-m4,Koike_Kaneko_Okazaki_2024, info14100522}.
Such systems often achieve promising performance on in-domain datasets, but may not generalize to out-of-domain data~\cite{wang-etal-2024-m4}. %inputs~\cite{wang-etal-2024-m4}.
This suggests that MGT detection systems may be more apt for some data than for others.
%This raises a question of whether %current 
%MGT detection systems are more apt for some data than for others.
%This suggests that detectors may be more apt for detecting MGT with certain stylistic tendencies.
Here, we audit %investigate the limitations of 
two state-of-the-art MGT detection methods by 
%Specifically, we subsample 
subsampling their evaluation datasets using linguistic features and readability measures and compare model performances on these subsets.
%compute linguistic features and readability measures to subsample evaluation datasets and we compare existing models' performance on these subsets.

% The NLP community tackled these challenges by developing classifiers to detect machine-generated text (MGT). These classifiers have reported promising accuracies on in-domain datasets, but have shown signs of lack of generalizability when evaluated on out-of-domain text samples. Thus, questions on the classifier's sensitivity to domain-specific writing styles and artifacts become crucial.
% LINGUISTIC FEATURES + MODELS
In an effort to investigate the limitations of MGT detectors, 
%detecting MGT, 
we evaluate the sensitivity of current detectors to stylistic variations and text complexity. 
Specifically, 
%To do so, 
we %evaluate the performance of 
examine two categories of detection systems trained on different domains, and categorize their test sets using linguistic features, and metrics for lexical composition, sophistication, and diversity, and readability. %\zee{How are lexical diversity and sophistication different?, diversity could be frequency metrics and sophistication is difficulty level}
%we evaluate the performance of two categories of detectors from varying domains and employ linguistic feature extraction techniques to categorize test set samples based on their lexical composition, readability, lexical diversity, and sophistication. 
We then evaluate model performances on different ranges for each feature category and report performances for each subset.\looseness=-1
%We then subsample the test set based on varying threshold ranges of each generated feature and compute performance for each subset, which enables us to identify a model's over-reliance on basic stylistic features.

We find that the classifiers are highly sensitive to the distribution of part-of-speech classes, %certain 
%word classes, 
e.g., %verbs and 
adverbs, to stylistic features, %writing style, including 
%such as 
e.g., average sentence length, % and readability, 
and to surface-level artefacts, e.g., punctuation. 
For instance, we find that the F1-score of detectors drops from $0.4\rightarrow0.0$ and $0.6\rightarrow0.3$ for different ratios of adverbs in human-written and machine-generated text.
%For instance, examining performance for different ratios of adverbs in human-written and machine-generated text, the accuracy %of one system 
%drops from 0.4 to 0.0, and 0.6 to 0.3 for each class.
%For instance, by controlling the ratio of adverbs in human-written and MGT, the performance for both classes drops substantially: from 0.4 to 0.0, and from 0.6 to 0.3, respectively.
%Analyzing the feature importance, we find that classifiers over-rely on surface-level artefacts (e.g., punctuation). 
Our findings %indicate 
suggest that performance of detectors across domains and styles is likely over-estimated. 
%while detectors may work well in certain domains, their performance could be over-estimated.
We therefore call for care in using such tools for critical societal functions, e.g., plagiarism detection in education, and recommend that future work attends to linguistic and stylistic artefacts in benchmark datasets.
%Thus, we argue that 
%We therefore argue that 
%we should be careful when including such tools in critical societal setups, such as plagiarism detection in education.

% FINDINGS
%Our findings illustrate that classifiers are highly sensitive to the distributionwof verbs, adverbs, average sentence length, and the readability of a passage. The presence of adverbs (0.09 < adverb ratio <= 0.14) degrades accuracy, with F1-scores dropping from 0.4 to 0 for human-written and from 0.6 to 0.3 for machine-generated text. Longer sentences increase the F1-score for detecting machine-generated text from 0.64 to 0.88 but reduce it for human-written passages from 0.45 to 0. The detectors are also susceptible to misclassify easy-to-read texts while illustrate high performance for complex texts. Our analysis also demonstrate over-reliance on surface-level artifacts such as punctuation marks (period and comma) and white-spaces. 

\begin{table*}[h!]
\centering
\resizebox{\linewidth}{!}{ % Resize table to fit within line width
\begin{tabular}{cccccc}
\hline
\textbf{Model} & \textbf{Training Data%Trained
} & \textbf{Evaluation Data} & \textbf{Macro F1-Score} & \textbf{Drop (\%)} \\ \hline
\multirow{8}{*}{LR-GLTR} & ArXiv (C-GPT \& GPT-3.5) & ArXiv (C-GPT) & 0.95 & - \\ \cline{2-5}
& ArXiv (C-GPT \& GPT-3.5) & ArXiv (CO) & 0.92 & \(\downarrow 3.16\%\) \\ \cline{2-5}
& ArXiv (C-GPT \& GPT-3.5) & ArXiv (GPT-3.5) & 0.79 & \(\downarrow 16.84\%\) \\ \cline{2-5}
& ArXiv (C-GPT \& GPT-3.5) & OUTFOX (C-GPT) & 0.60 & \(\downarrow 36.84\%\) \\ \cline{2-5}
& ArXiv (C-GPT \& GPT-3.5) & IDMGSP (C-GPT, GA) & 0.53 & \(\downarrow 42.11\%\) \\ \cline{2-5}
& OUTFOX (C-GPT) & OUTFOX (C-GPT) & 0.91 & - \\ \cline{2-5}
& OUTFOX (C-GPT) & IDMGSP (C-GPT, GA) & 0.53 & \(\downarrow 41.76\%\) \\ \hline
\multirow{2}{*}{RoBERTa} & ArXiv (C-GPT \& GPT-3.5) & ArXiv (C-GPT) & 0.99 & - \\ \cline{2-5}
& ArXiv (C-GPT \& GPT-3.5) & IDMGSP (C-GPT, GA) & 0.33 & \(\downarrow 66.00\%\) \\ \hline
\multirow{2}{*}{GPT-Large} & OPEN AI Detector & GPT2 Generations & 0.95 & - \\ \cline{2-5}
& OPEN AI Detector & IDMGSP (C-GPT, GA) & 0.80 & \(\downarrow 15.00\%\) \\ \hline
\multirow{1}{*}{Llama-3.1-8B} & Zero Shot & IDMGSP (C-GPT, GA) & 0.50 & - \\ \hline
\end{tabular}
}
\caption{Comparison of in-domain and out-of-domain performance of detectors. The ``Drop'' column represets the decrease in F1-score from the in-domain configuration to the out-of-domain configuration.}
\label{tab:performance_comparison}
\end{table*}

\section{Related Work}

Prior work has sought to detect MGT by %relying 
using on feature-based~\cite[e.g.,][]{frohling2021feature, prova2024detecting} and neural network-based~\cite[e.g.,][]{gaggar2023machinegenerated} methods, reporting over 80\% and 90\% accuracy, respectively.
%, some relying on feature-based methods~\cite[e.g.,][]{frohling2021feature, prova2024detecting} while others used neural network-based approaches~\cite{wang-etal-2024-m4, gaggar2023machinegenerated}.
%Both have reported high performance, with the former achieving more than 80\% accuracy~\cite{prova2024detecting,frohling2021feature}, and the latter over 90\%~\cite{wang-etal-2024-m4, prova2024detecting}.
%Both approaches have reported high performances, with feature-based methods reporting accuracies above 80\%~\cite{prova2024detecting,frohling2021feature} and LLM-based methods over 90\%~\cite{wang-etal-2024-m4, prova2024detecting}.
This body of work has primarily used three feature types for MGT detection:
%Three %different 
%feature types are commonly used for MGT detection:
%Often, three different types of features are relied upon for MGT:
statistical distributions (e.g., log-likelihood)~\cite[e.g.,][]{gehrmann-etal-2019-gltr}, features obtained from fact-checking methods~\cite[e.g.,][]{wang-etal-2024-m4}, and linguistic features~\cite[e.g.,][]{tang2023science}.
%
%Several works have been conducted to detect machine-generated work, with some opting for feature-based detection techniques \cite{frohling2021feature, prova2024detecting} and others utilizing large language neural networks \cite{wang-etal-2024-m4, gaggar2023machinegenerated}. Both approaches have been \emph{somewhat} successful, with classical techniques reporting accuracy above 80\% \cite{prova2024detecting,frohling2021feature} and large neural networks above 90\% \cite{wang-etal-2024-m4, prova2024detecting}. Three types of features are generally used in MGT detection: statistical distributions (e.g., log-likelihood) \cite{mitchell2023detectgpt, gehrmann-etal-2019-gltr}, fact verification features, and linguistic features \cite{tang2023science}. 
%

Other work has proposed zero-shot approaches to MGT detection: 
For instance~\newcite{mitchell2023detectgpt} rely on log-probabilities from the generating model and random perturbation of the text from another generic LLM; and \citet{guo2023authentigpt} use a black-box LLM to denoise input text with artificially added noise, and then semantically compare the denoised and original text.
Yet other work has examined the use of watermarks for MGT as a mechanism for detecting MGT. 
For example, \citet{pmlr-v202-kirchenbauer23a} propose using soft-constraints through \textit{green} and \textit{red} lists of vocabulary to include or exclude from MGT. %thus reducing the task of detecting MGT to detecting the presence of a watermark.
%
%followed by semantic comparisons between the denoised text and the original text for MGT detection~\cite{guo2023authentigpt}. 

%DetectGPT \cite{mitchell2023detectgpt} proposed a zero-shot approach for MGT detection that does not require training a separate classifier using log probabilities computed by the model of interest and random perturbations of the passage from another generic pre-trained language model \cite{mitchell2023detectgpt}. 
%AuthentiGPT \cite{guo2023authentigpt} proposed a black-box LLM to denoise input text with artificially added noise, followed by the semantic comparison of the denoised text and the original text for MGT detection. 
%
Recent work has also conducted comprehensive analyses of MGT detection methods and resources~\cite[e.g.,][]{tang2023science,jawahar-etal-2022-automatic, mitchell2023detectgpt, guo2023authentigpt}.
\citet{tang2023science}, for example, highlight for the need %for developing 
measures for evaluating MGT detection systems. 
They argue that current evaluation measures (e.g., AUC and accuracy) are limited for security analysis by only considering the average instance and are limiting for security analysis.
% \zee{This reads like a sentence stub. The average of what? It was commented out. fixed} 
Similarly, watermarks for MGT have been called into question, with \citet{zhang2024watermarks} arguing that ``strong watermarking of generative models is impossible.''
%Watermarking has also been critically examined in the literature. 
%\citet{zhang2024watermarks} examine and break existing watermarking techniques, and argue that ``strong watermarking of generative models is impossible.''

%Some work~\cite[e.g.,][]{tang2023science,jawahar-etal-2022-automatic, mitchell2023detectgpt, guo2023authentigpt} has conducted comprehensive analysis of MGT detection methods and resources. 
%\citet{tang2023science} emphasized the importance of developing measures for evaluating MGT detection. 
%They discussed in detail how current evaluation measures (e.g., AUC and accuracy) only consider the average instance and are limiting for security analysis. 
%Other work has focused on presenting datasets for MGT~\cite[e.g.,][]{wang-etal-2024-m4,Koike_Kaneko_Okazaki_2024,info14100522} which typically contain human-written text from a domain, and the outputs of LLMs prompted to generate text conditioned on partial information from the human-written text.
%For example, in one segment of the dataset presented by \citet{wang-etal-2024-m4}, they use titles from ArXiv papers to generate abstracts using five different LLMs to contrast human-written abstracts.

Research has therefore attempted to develop datasets for detecting MGT~\cite[e.g.,][]{wang-etal-2024-m4,Koike_Kaneko_Okazaki_2024,info14100522,gpt-2dataset}. 
Such datasets typically contain human-written texts for given domains, and the generated outputs of LLMs that have been conditioned on partial information from the human written texts~\cite[e.g.,][]{wang-etal-2024-m4,guo2023close,he2023mgtbench}. 

\section{Experiments}
We evaluate MGT detection using three datasets and four classifiers, and use linguistic features for analysis. 
Here, we describe our experimental setup.
%To evaluate classifiers for MGT detection, we use three datasets, classifiers, and linguistic features for analysis.
%In this section, we describe the artifacts and features we use for our experiments.
%In this section, we describe the datasets, classifiers, and features used for our experiments.
%RIn this section, we describe the datasets and features we rely on for our analysis of weaknesses of MGT detection systems.

\subsection{Data}
We conduct experiments using three datasets: M4, OUTFOX, and IDMGSP.

%We use the following three datasets for our experiments:

\paragraph{M4} 

The M4 dataset~\cite{wang-etal-2024-m4} consists of ~147K human-written texts across data sources and languages, paired with human-written and MGTs generated by several LLMs. % including Wikipedia, Reddit, and ArXiv in the English subset of the dataset.
For our experiments, we use the English subset of the M4 dataset, which consists of 102K human-written texts, sourced from Wikipedia, WikiHow, Reddit, ArXiv and PeerRead, and outputs from GPT-4, ChatGPT and text-davinci-003 (henceforth GPT-3.5).

%We use the English subset of titles and abstracts from ArXiv and texts generated using text-davinci-003 (henceforth GPT-3.5), GPT-4, and ChatGPT.
%It pairs the human-written texts with texts generated bywseveral LLMs, including text-davinci-003 (henceforth GPT-3.5), GPT-4, and ChatGPT. 
%The dataset includes human-written texts from sources such as Wikipedia, WikiHow, Reddit, ArXiv, and PeerRead for English. 
%Machine-generated texts were created using models including GPT-4, ChatGPT, and GPT-3.5 (text-davinci-003). 
%The English subset of the dataset contains %approximately 147,000 human-machine parallel texts, with 
%102K human-written texts. %in English and 45,000 in other languages. 

% TODO: Review portion of English M4 used
% For our experiments, we use the generated ArXiv-like abstracts from ChatGPT and GPT-3.5.

\paragraph{OUTFOX} The OUTFOX dataset~\cite{Koike_Kaneko_Okazaki_2024} consists of 15K %,400 
triplets of essay problem statements, student-written essays, and machine-generated essays.
We use human-written and ChatGPT-generated essays for training, and GPT-3.5-generated essays for testing. 

%We use the human-written and ChatGPT-generated essays for training and we use the GPT-3.5-generated essays for testing.

% \zee{Insert number here} --> RESOLVED
\paragraph{IDMGSP} The IDMGSP dataset~\cite{info14100522} contains 4K %,000 
human-written and 4K %,000 
machine-generated (SCIgen, GPT-2, ChatGPT, and Galactica) scientific papers. 
We restrict our analysis to abstracts of scientific papers because they are similar in length, which allows for a fair comparative evaluation across different samples.

% We create a test set of abstracts for evaluating our detectors.
%In our work, we create a test set of abstracts to evaluate our detectors.

% ==========================
\begin{figure*}[h!]
    \centering
\includegraphics[width=\columnwidth * 2]{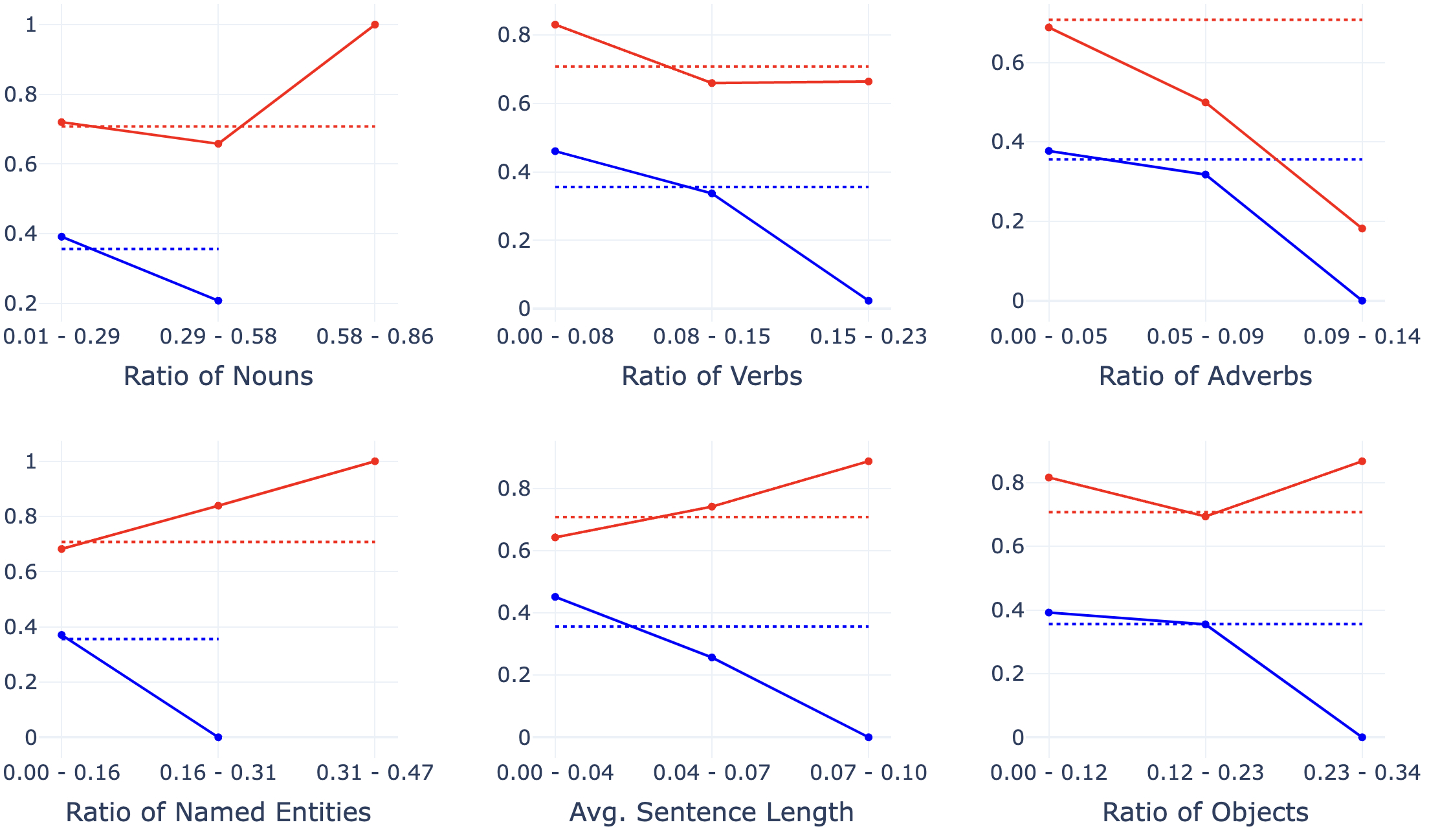}
    \caption{F1-scores for %the 
    LR-GLTR %classifier 
    (trained on M4) %and tested on 
    for IDMGSP. % across different data segments. 
    {\color{red}Red indicates machine-generated},
    %Colors indicate {\color{red}machine-generated} 
    {\color{blue}blue human-written} data, and dashed lines indicate baselines.}
    %for machine-generated (red) and human-written (blue) data. 
%    Dashed lines indicate the baseline performance for each class.\looseness=-1}
%    This plot illustrates the F1-scores of LR+GLTR classifier trained on ArXiv (ChatGPT \& Davinci) and evaluated on IDMGSP (abstracts) test set sub-samples across different feature ranges for machine-generated (red) and human written (blue) classes.}
    \label{fig:fmeasure_drops}
    %\vspace{-2em}
\end{figure*}
% ===========================

\subsection{Machine-Generated Text Classifiers}
We evaluate neural and feature-based methods for MGT detection.
The neural %detection 
methods rely on 
%Neural detection methods typically rely on 
fine-tuning %pre-trained 
LLMs, while the feature-based methods rely on %either hand-crafted or 
machine-generated features for MGT detection. %\zee{Changed metric-based to feature-based because of what's written below but CHECK.}
%use the probabilities of existing generative models.\zee{This paragraph uses metric-based, but the entire metric-based approach doesn't make sense.}
%We evaluate the performance of two categories of detectors: neural and metric-based. Neural detectors are usually fine-tuned on top of an encoder style pre-trained language model On the other hand, metric-based detectors like LR+GLTR typically rely the probabilities of an existing generative model. We tested the following detectors:

\paragraph{Neural Methods}
We use two fine-tuned version of RoBERTa~\cite{Liu2019RoBERTaAR}: the \textbf{OpenAI Detoctor}, a RoBERTa-Large model fine-tuned on GPT-2-generated texts, and \textbf{RoBERTa-M4}, a RoBERTa-base model fine-tuned on the M4 dataset following \citet{solaiman2019release} and \citet{wang-etal-2024-m4}.
%We fine-tuned two versions  of RoBERTa, a pre-trained encoder language model~\cite{Liu2019RoBERTaAR}, to predict whether a text is human-written and machine-generated.
%%
%Specifically, we use the \textbf{OpenAI Detector}, a RoBERTa-Large detector released by OpenAI, trained on text generated using GPT-2 \cite{Radford2019LanguageMA}, and \textbf{RoBERTa-M4}, a RoBERTa-base model fine-tuned on %our subsample of 
%the M4 dataset following \citet{solaiman2019release} and \citet{wang-etal-2024-m4}, respectively.

\paragraph{Feature-Based Methods}
Following \citet{wang-etal-2024-m4}, we use \textbf{LR-GLTR}, a logistic regression model trained using 14 features from \citet{gehrmann-etal-2019-gltr}.
The model uses two sets of features: The number of tokens within the top-\{10, 100, 1000, 1000+\} %, top-100, top-1000, and 1000+ 
ranks from a LM's %language model's 
predicted probability distribution (4 features); the probability distribution for a given word divided by the maximum probability for any word in the same position over 10 bins ranging from 0.0 to 1.0 (10 features).
%
%We followed \citet{wang-etal-2024-m4} and used \textbf{LR-GLTR}, a logistic regression model trained on 14 features from \citet{gehrmann-etal-2019-gltr}.
%This model uses two sets of features: the number of tokens within the top-10, top-100, top-1000, and 1000+ ranks from a language model's predicted probability distribution (4 features); and %\zee{Osama, can you describe it after this comment, so it fits with the rest of the sentence?} 
%the probability distribution for a given word divided by the maximum probability for any word in the same position over 10 bins ranging from 0.0 to 1.0 (10 features).
%the Frac(p) distribution over 10 bins ranging from 0.0 to 1.0 (10 features).
%Frac(p) describes the fraction of probability for the actual word divided by the maximum probability of any word at this position.
We train one instance of this model on a subset of M4, and another instance on subset of OUTFOX.
%We train two instances of this model: one on a subset of the M4 dataset, and the other on a subset of the OUTFOX dataset.

\paragraph{Zero-Shot Prediction Methods}
We use \textbf{Llama-3.1-8b} for zero-shot classification of machine-generated versus human-written text.
%We employed \textbf{Llama-3.1-8b} in a zero-shot setting to detect human versus machine-generated text. The model uses its pre-trained language capabilities to perform the classification without task-specific fine-tuning.

\subsection{Linguistic Features for Analysis} 
We extract linguistic and extra-linguistic features for analysis. 
Specifically, we extract \textbf{Part-of-Speech (POS)} tags and named entities using spaCy~\cite{honnibalSpaCyNaturalLanguage2017}.
We then compute \textbf{average sentence length}, and the \textbf{ratio of nouns, verbs, adverbs, named entities, and objects} (direct or prepositional) in a text, i.e., the number of occurrences divided by the total number of tokens in the text.
We also compute the \textbf{Flesch Reading Ease} score ~\cite{Flesch_New_1948a} to assess the reading difficulty of a text.
This metric is computed using sentence length, syllable density, and word familiarity.
Finally, we compute the lexical diversity of texts, i.e., the variety and range of words used.
Specifically, we compute \textbf{Hapax (Legomena)}, the number of words that occur only once in a text, and \textbf{Dihapax (Dis Legomena)}, the number of words that occur twice in a text.
Hapax and Dihapax help illustrate the richness of the vocabulary used.

\section{Analysis}
%\Cref{tab:performance_comparison} depicts performance drops for all models and held-out evaluation datasets.
%In this section, 
Here, we investigate the sources of classification errors
%examine some sources of classification error 
(see \Cref{tab:performance_comparison} for impacts of domain shifts).

\begin{figure*}[h!]
    \centering
    \includegraphics[width=\columnwidth * 2]{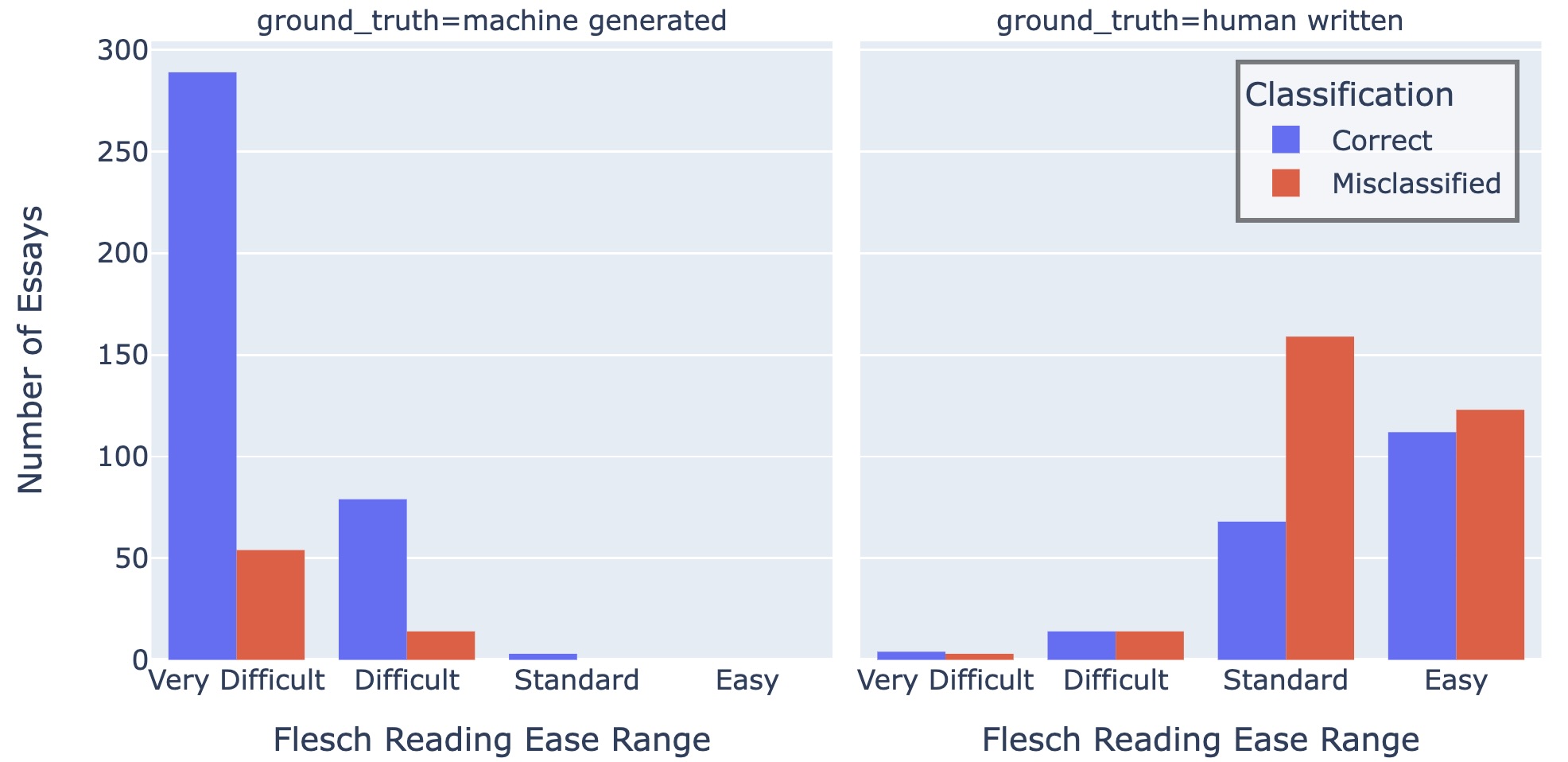}
    \caption{%Classification 
    %Impact of 
    Flesch Reading Ease analysis of  %levels on classification using 
    %the performance of 
    LR-GLTR (ArXiv) evaluated on OUTFOX (essays).
    %Performance of LR-GLTR (ArXiv) %trained on ArXiv 
    %evaluated on OUTFOX (essays)
    %and tested on OUTFOX (essays) %based 
    %on Flesch Reading Ease levels. %score.
    }
    \label{fig:missclassification_flesch}
\end{figure*}

\paragraph{Impact of Surface Form Linguistic Features}
Ideally, a MGT detector would not overfit to linguistic surface features, however, we find that the \textbf{LR-GLTR} model significantly overfits to such features (see \Cref{fig:fmeasure_drops}).\footnote{
Only machine-generated texts appear in the (0.58–0.86] noun ratio range.}
For instance, we see that the model performance for both human-written and machine-generated text drops to near zero as the ratio of adverbs increases. 
Moreover, as we the ratio of named entities, objects, and the average sentence lengths increase, the model performances drop to zero for human-written texts, while obtaining near perfect scores for machine generated texts. 
That is, beyond a given ratio of linguistic surface form items, and average sentence length, the model loses the capability to identify human-written text.

\subsection{Impact of Readability}
Turning our attention to readability (see \Cref{fig:missclassification_flesch}), we find that LR-GLTR has high accuracy for very difficult passages, as it has very few classification errors.
However, the model's ability to correctly classify MGT decreases as the reading difficulty decreases. 
For human-written text, a different pattern emerges: The model struggles to correctly classify human-written texts regardless of text difficulty.
%In contrast, the classifier struggles to correctly predict human-written texts that are easy, often mistaking them for machine-generated.

% Yet when the human-written texts are rated on the more difficult end of the reading ease scale, the classifier improves its performances.
% \zee{This last sentence does not seem to be true looking at the figure. It seems like the classifier gets to a random classifier}
%However, as the readability improves, the classifier struggles more to recognize that the text was machine-generated.

%For machine-generated essays, the detection classifier illustrates high accuracy (shown in \Cref{fig:missclassification_flesch}) when the passages are very difficult to read. It correctly classifies machine-written texts with high accuracy and few mistakes. However, as the readability improves, the classifier struggles more to recognize that the text was machine-generated. On the flip side, the classifier has trouble with human-written essays that are standard or easy to read, often misclassifying them. However, when human-written passages are very difficult or difficult in terms of readability, the classifier is better able to correctly identify them as human-written. 
% Essentially, the classifier is good at spotting machine-written texts that are very complex, and human texts that are very complex. Its accuracy decreases when dealing with texts on either side that are simple and easy to read, whether machine or human-generated. 

\subsection{Impact of Punctuation Marks}
\begin{figure}[]
    \centering
\includegraphics[width=8cm]{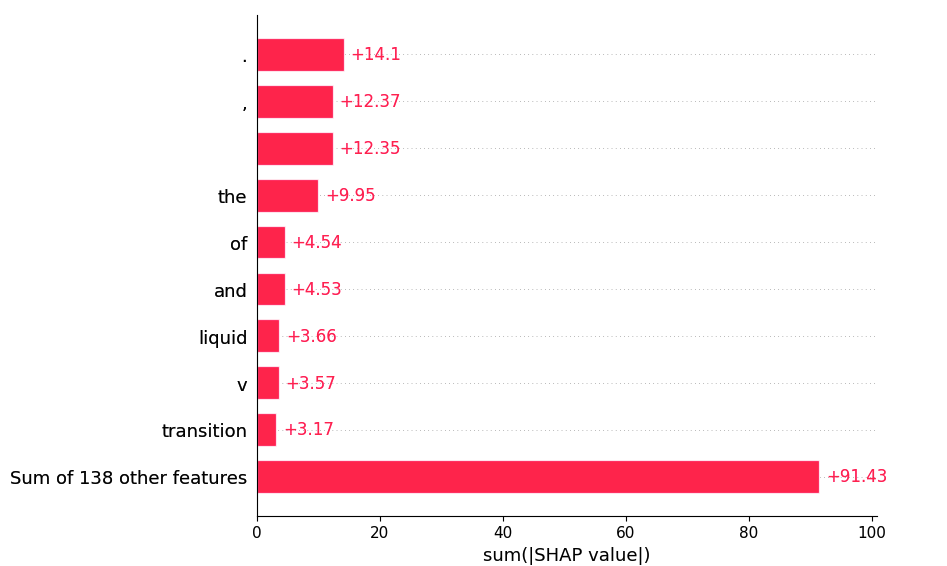}
    \caption{Feature importance for RoBERTa trained on ArXiv using SHAP values.}
    \label{fig:roberta_feature_importance}
\end{figure}

To investigate RoBERTa's sizable performance drop (see \Cref{tab:performance_comparison}) on out-of-domain evaluation sets, we use Shapley Additive Explanations (SHAP)~\cite{Lundberg_Unified_2017}, which quantifies the impact of a given feature on a model's performance.
We find that punctuation marks and whitespace (see \Cref{fig:roberta_feature_importance}) are among the most important features.
Such over-reliance on punctuation suggests that the model is overfitting and therefore not learning general features of MGT.
%The heavy reliance on such features, and the associated out-of-domain performance drop, indicates that the model may be overfitting to the training data instead of learning general features of MGT.
%model may not be learning general features of machine-generated text, but rather that is overfitting to the training dataset.

Across both RoBERTa and LR-GLTR models, it appears that surface level features are highly influential for classifier performance.
In turn, this suggests that simple adversarial attacks such as changing the ratio of nouns, adverbs, or changing punctuation can render these models ineffective.

% ==========================
% ===========================

%Investigating RoBERTa's large performance gap on out-of-domain test sets, we measure the importance of each feature using SHAP (SHapley Additive exPlanations) which quantifies the impact of each feature on the model's output. As shown in \Cref{fig:roberta_feature_importance}, punctuation marks (period and comma) as well as whitespaces are among the most important features. 
%If a model heavily relies on punctuation marks and whitespaces, it may be overfitting to specific surface-level patterns or domain-specific artifacts presented in the training data rather than capturing more nuanced linguistic features, leading to poor generalization. 
%Such models would be susceptible to adversarial attacks as slight changes in punctuation usage can render the model innefective.

\subsection{Impact of Lexical Diversity}
\begin{figure}[tbh]
    \centering
\includegraphics[width=8cm]{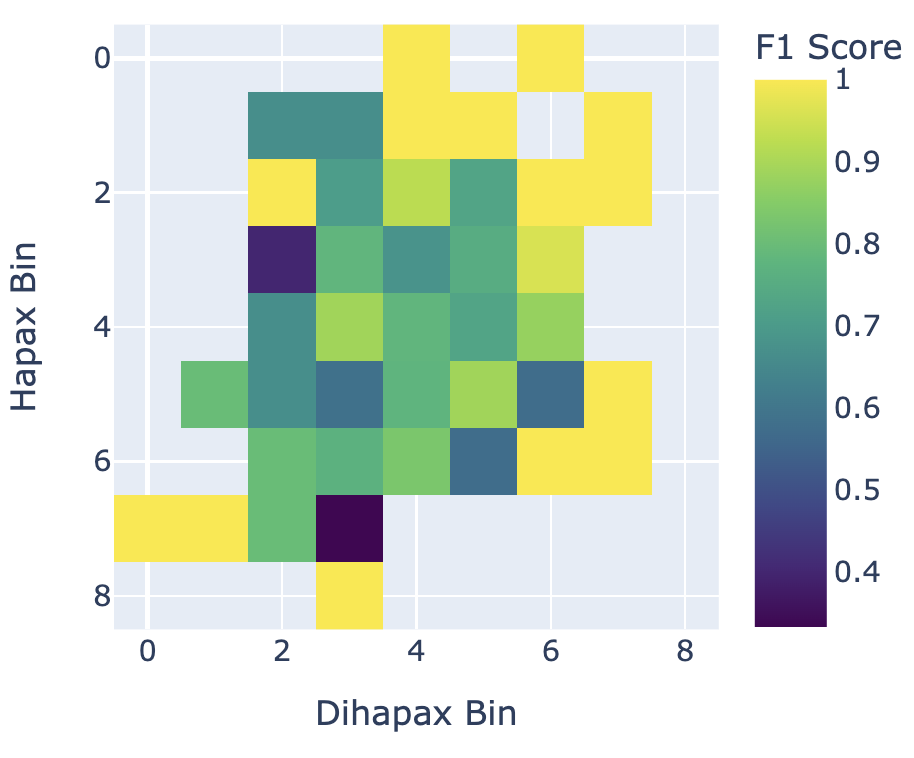}
    \caption{%Heatmap of 
    F1 scores for LR-GLTR trained on ArXiv (ChatGPT \& GPT-3.5) and tested on ArXiv (GPT-3.5) across %different 
    bins of hapax and dihapax features.}
    \label{fig:heatmap_f1_hapax_dipax}
\end{figure}

Considering lexical diversity (see \Cref{fig:heatmap_f1_hapax_dipax}), the classifier performs best when detecting texts with a narrow vocabulary (low Hapax bins) and specific repetition patterns (high Dihapax bins).
For example, the model has high performance for Hapax bins 0 and 1, when combined with Dihapax bins 6 and 7.
In contrast, the model struggles with texts that have a rich and varied vocabulary (high Hapax) and certain combinations of repetitions.

% ==========================
% ===========================

\subsection{Impact of Named Entities}
\begin{figure}[h!]
    \centering
\includegraphics[width=\columnwidth]{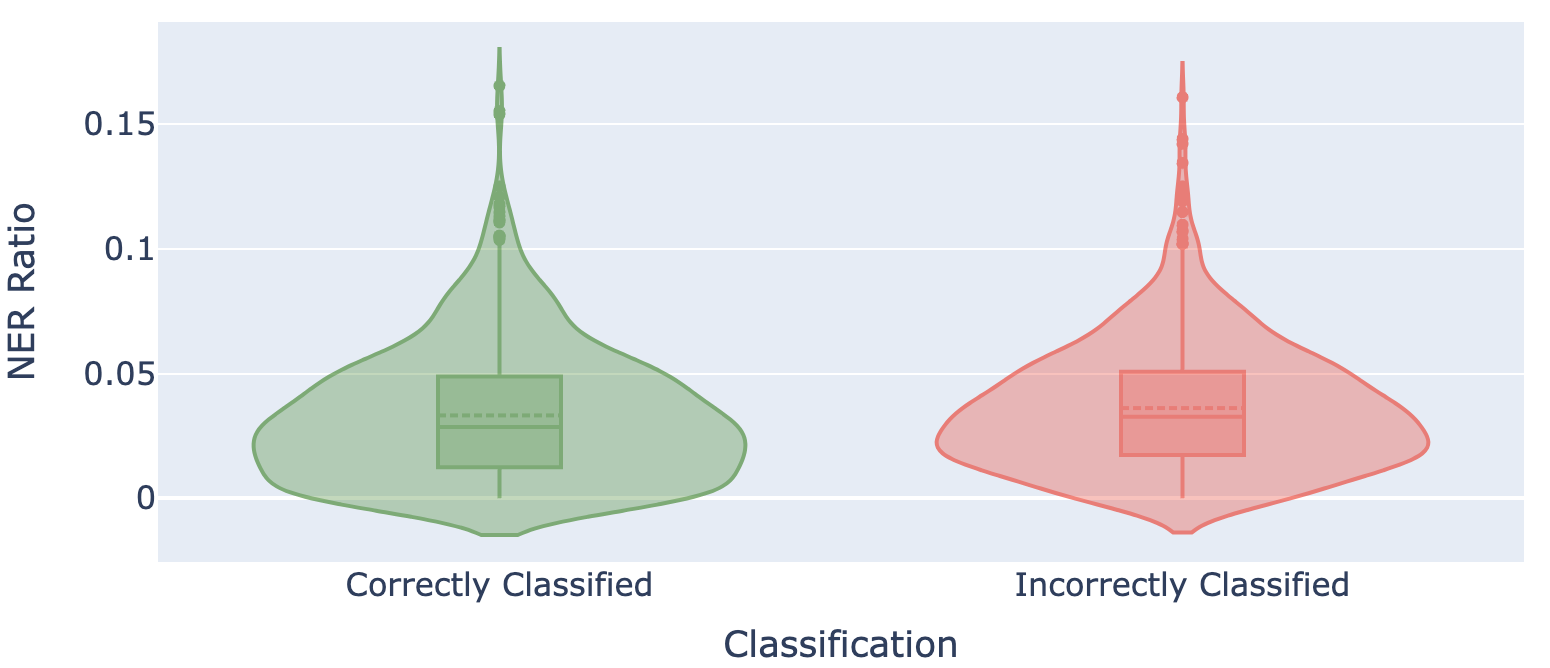}
    \caption{%Distribution of 
    NER ratio for %Correctly and Incorrectly 
    classified abstracts.}
    \label{fig:missclassification_ner}
\end{figure}

We conducted an analysis to evaluate the impact of the Named Entity Recognition (NER) ratio---defined as the number of named entities relative to the total token count---on the zero-shot classification performance of scientific abstracts using Llama-3.1-8B (see \Cref{fig:missclassification_ner} and \Cref{fig:missclassification_ner_hist}). % zero-shot model for scientific abstracts. 
We find that correctly classified abstracts have a broader range of NER ratios, while incorrectly classified abstracts concentrate named entities in the lower rations.
%have higher concentration of  in the lower ratios. 
That is, the model is more prone to misclassify abstracts with fewer named entities.
We test this finding by computing Welch's T-test for NER ratios of correctly and incorrectly classified samples.
%We compute Welch's T-test for the named entity ratios for correctly and incorrectly classified samples. 
We find that the hypothesis---that the two distributions have the same mean---to be rejected %\footnote{
($t=-2.289$, \textit{degrees of freedom} $=1680$, \textit{critical} $t=1.96$, \textit{p-value}$=0.022$), which indicates a statistically significant difference in the means ($p < 0.05$). 
The t-test suggests that the mean NER ratio is lower for incorrectly classified abstracts, i.e., that performance decreases as NER ratio decreases.

%A T-test was performed to compare the NER ratios between correctly and incorrectly classified abstracts, resulting in a T-statistic of -2.289 and a P-value of 0.022, which indicates a statistically significant difference in the means ($p < 0.05$). 
%The negative T-statistic suggests that the mean NER ratio is lower for incorrectly classified abstracts, meaning that the model's performance decreases when the NER ratio decreases. 

%\Cref{fig:missclassification_ner} illustrates the distribution of NER ratios for correctly and incorrectly classified abstracts by the LLAMA 3.1 8B zero-shot model. 
%The violin plot shows that correctly classified abstracts have a broader range of NER ratios, while incorrectly classified abstracts are concentrated within a narrower range with higher density around lower NER ratios. 
%This suggests that the model tends to misclassify abstracts with fewer named entities, which aligns with the statistical test results.

\section{Risk of Deployment}
% Deploying MGT detectors %, such as the ones evaluated above, 
% comes with risks of reliability and fairness. %pertaining to reliability and fairness. 
% For instance, an MGT classifier's over-reliance on surface-level features, e.g., writing style and punctuation marks makes it vulnerable to adversarial attacks. 
% Adversaries can %intentionally 
% alter these features to cause misclassification. 
% Further, over-fitting to a specific writing style can lead to unfair misclassification of subgroups with a specific writing style resulting in disparate outcomes.
% Further, as new models are released, the writing style and fluency of MGT changes, which can render existing detectors ineffective.
% Similarly, our results suggest that MGT detectors are sensitive domain shift, and therefore require careful consideration to match the domain of training data and application domain.
% We recommend that future work in developing datasets for MGT detection attend to surface-level features that models may overfit to.

Deploying MGT detectors, such as the ones evaluated above, comes with risks of reliability and fairness. The primary risks associated with deploying such systems are detailed below.

\paragraph{Adversarial attacks}
The classifier's over-reliance on surface-level features such as writing style, punctuation marks, and whitespace makes it vulnerable to adversarial attacks. Adversaries can intentionally alter these features to cause misclassification.

\paragraph{Bias and fairness}
Over-fitting to a specific writing style can lead to unfair misclassification of subgroups with a specific writing style. This can result in biased outcomes, particularly against individuals from different cultural, educational, or linguistic backgrounds—especially in contexts that encourage the use of richer vocabulary and longer sentences.

\paragraph{Data drift due to new LLMs}
The writing style and fluency of MGT change upon the release of new models, which can cause data drift, potentially rendering a classifier ineffective. As a result, classifiers may need regular retraining to accurately detect machine-generated text from newer models.

\paragraph{Domain shift sensitivity}
The results above indicate that although a classifier performs well within the same domain, it may be sensitive to domain shift. This sensitivity could limit a classifier's applicability and deployment in diverse settings.

% \begin{itemize}
%    \item \textbf{Adversarial attacks:} The classifier's over-reliance on surface-level feature such as writing style, punctuation marks, and white space makes it vulnerable to adversarial attacks. Adversaries can intentionally alter these features to cause misclassification.
%    \item \textbf{Bias and fairness:} Overfitting to a specific writing style can lead to unfair misclassification of subgroups with a specific writing style. This can result in biased outcomes, particularly against individuals from different cultural, educational, or linguistic backgrounds, especially when some contexts encourage the use of richer vocabulary and longer sentences.
%    \item \textbf{New LLMs causing data drift:} the writing style and the fluency of MGT changes upon the release of new models, which can cause data drift, potentially rendering a classifier in-effective. 
%    \item \textbf{Domain shift sensitivity:} The results above indicate that although a classifier performs well within the same domain, it may be sensitive to domain shift. This sensitivity could limit a classifier's applicability and deployment in diverse settings.
% \end{itemize}

\section{Conclusion}
In this paper, we have examined the limitations of several classifiers for detecting machine-generated text by evaluating their sensitivity to stylistic variation across domains.
We find that classifiers show high sensitivity to certain linguistic features, e.g., the distribution of adverbs, sentence length, and readability of the text.
Moreover, we find that classifiers overfit to punctuation marks and whitespace.
Our results suggest that current datasets for MGT are not robust to stylistic or domain shifts, and are particularly weak when applied to simple writing, e.g., school assignments.
We therefore call for the further development of datasets of MGT and critical assessments of MGT detection systems with data from their particular domain of interest to avoid potential negative consequences of misclassification in critical domains.\looseness=-1

%we explored the limitations of several classifiers for detecting machine-generated text (MGT) by evaluating their sensitivity to stylistic variations across different domains. 
%Our findings illustrate that classifiers are highly sensitive to certain linguistic features, i.e., %such as 
%the distribution of adverbs, average sentence length, and readability of the text. 
%%For instance, longer sentences tend to improve the detection of machine-generated text while reducing the accuracy of detecting human-written passages. 
%%While the detectors show high performance for complex texts, they can struggle with easy-to-read passage.
%%The detectors also show high performance for complex texts while struggling with easy-to-read passages. 
%Moreover, we highlight the over-reliance of classifiers on punctuation marks and whitespace. 
%Our results indicate that applying MGT detection systems come with risks, particularly when applied to simple writing, e.g.,~school assignments.
%We therefore call for researchers and practitioners to critically evaluate MGT detection systems 
%with data from the particular domain in which they are to be applied, 
%to avoid the potential negative consequences of misclassification, in critical domains.

\section*{Limitations}
Below are some of our main limitation pertaining to availability of labelled data and the dynamic nature of LLM generation:
\begin{itemize}
    \item \textbf{Dataset limitations:} The datasets used in this paper do not represent the full spectrum of potential domains. This is caused by the limited availability of labeled MGT data.
    \item \textbf{Dynamic nature of LLMs:} We assumed that text generation by a given model are static. However, LLMs are regularly updated and may exhibit changes in their writing style and coherence. However, such changes will typically cause detectors to fail beyond what is described in this work, further emphasizing the need for more careful data analysis.
\end{itemize}

\section*{Ethical Considerations}
Our paper investigates the performance of models for the detection of machine-generated text and emphasizes the careful testing and precise reporting of the performance of such systems.
This is particularly important, as our examined models struggle on less complex texts, which can have downstream impact if such systems are deployed in educational settings.
In light of our findings, we stress the importance of critically evaluating systems for detecting machine-generated text within the domains a given model is to be deployed.

% \section*{Ethics Statement}
% Entries for the entire Anthology, followed by custom entries
\bibliography{anthology,custom}
\bibliographystyle{acl_natbib}

\appendix

\section{Distribution of NER Ratios}
Here, we include a histogram (complementing the violin plot in \Cref{fig:missclassification_ner}) to illustrate the distribution of the NER ratio for correctly and incorrectly classified abstracts. 
% ==========================
\begin{figure}[h!]
    \centering
    \includegraphics[width=8cm]{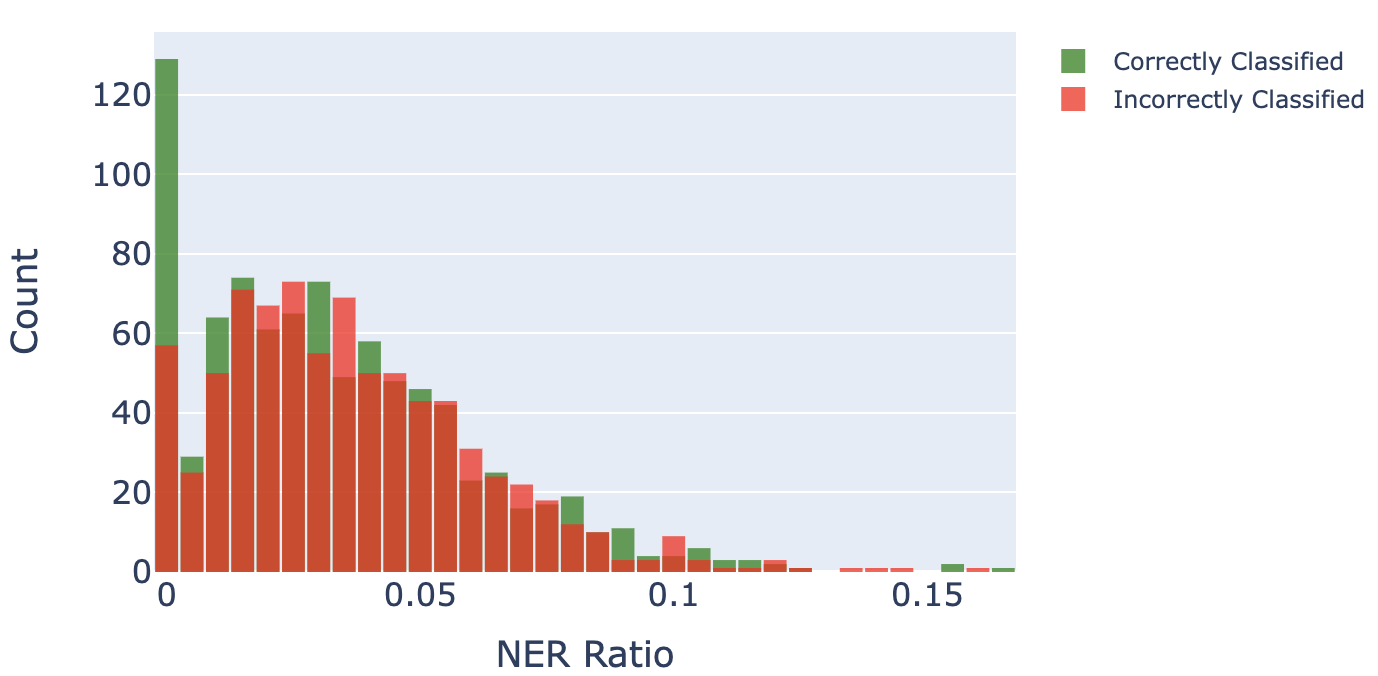}
    \caption{Distribution of NER Ratios (Histogram)}
    \label{fig:missclassification_ner_hist}
\end{figure}
% ===========================

\section{Part of Speech Analysis}
Here, we extend the part-of-speech (POS) sub-sampling evaluation with a broader coverage of models to explore properties influencing classifier performance across different models and domains. The evaluated models include zero-shot detection methods (LLama 3.1 8b and Binoculars \cite{hans2024spotting}) and variants of the LR-GLTR models.

\begin{figure}[h!]
    \centering
    \includegraphics[width=8cm]{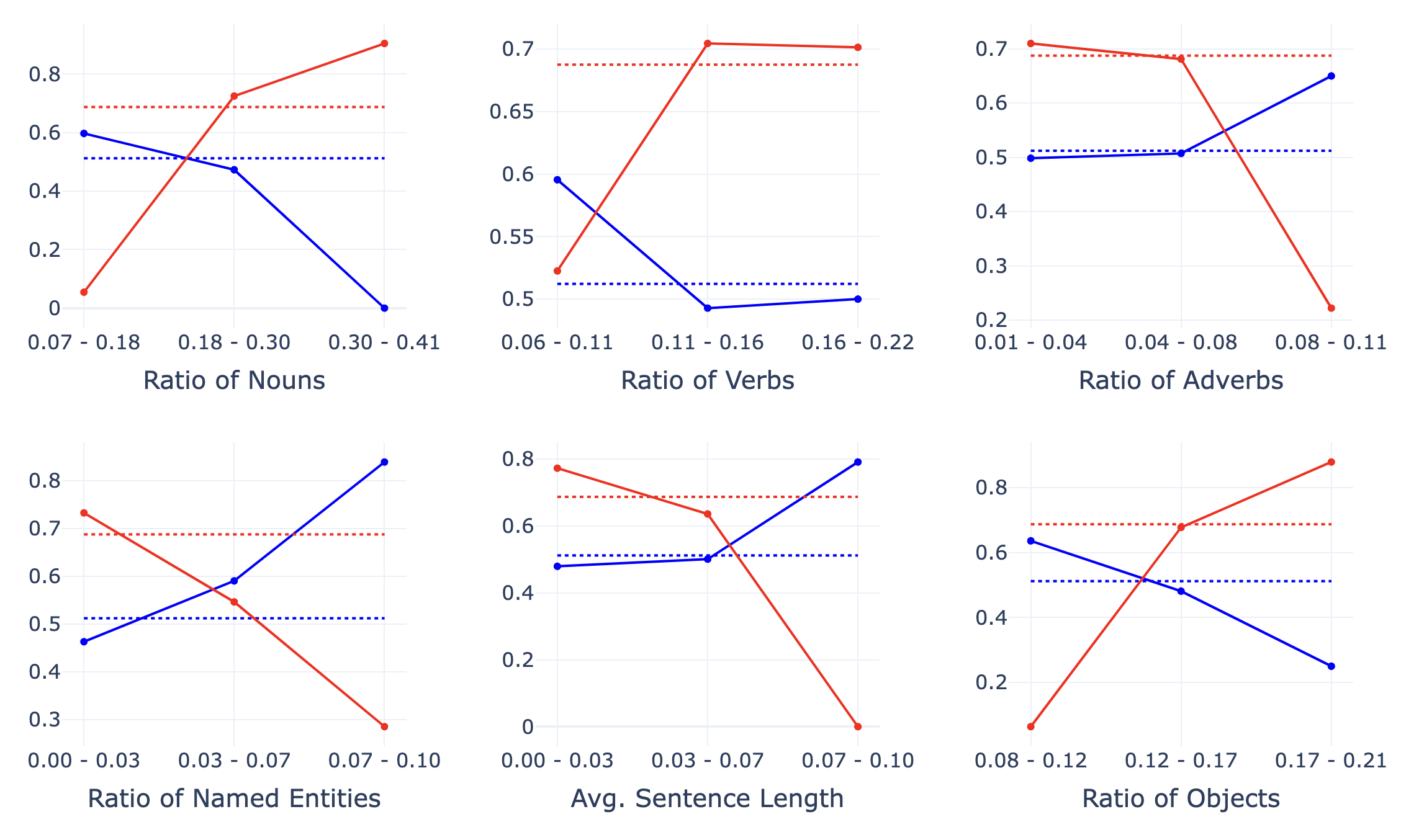}
    \caption{GLTR Logistic Regression: Train ArXiv, Test Essays (ChatGPT). {\color{red}Red indicates machine-generated},
    {\color{blue}blue human-written} data, and dashed lines indicate baselines.}
    \label{fig:gltr_train_arxiv_test_essays_chatgpt}
\end{figure}

\begin{figure}[h!]
    \centering
    \includegraphics[width=8cm]{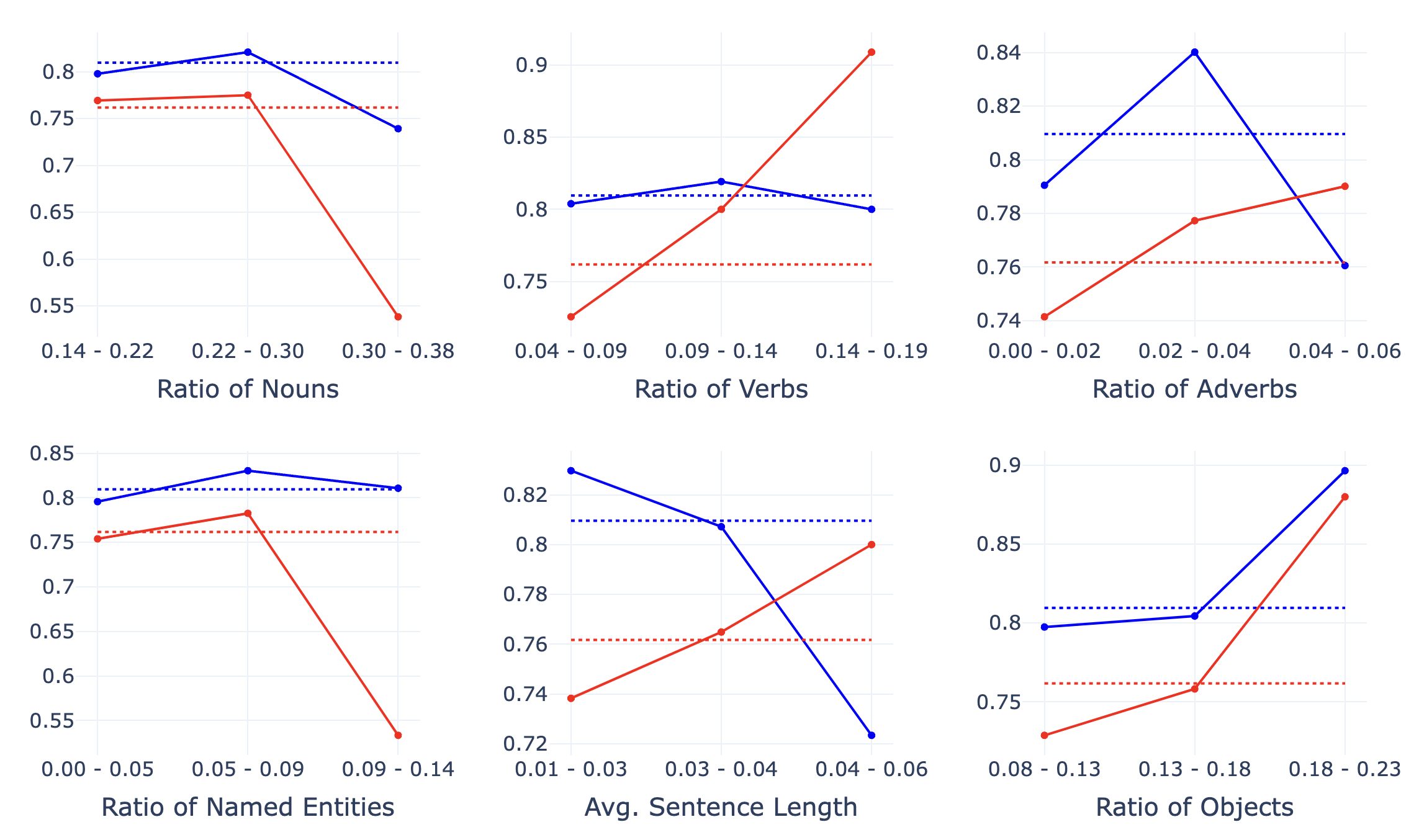}
    \caption{GLTR Logistic Regression: Train ArXiv, Test ArXiv (Davinci). {\color{red}Red indicates machine-generated},
    {\color{blue}blue human-written} data, and dashed lines indicate baselines.}
    \label{fig:gltr_train_arxiv_test_arxiv_davinci}
\end{figure}

The results indicate that POS features do not generalize to out-of-domain samples (as seen in \Cref{fig:gltr_train_arxiv_test_essays_chatgpt}) but retain F scores above 0.5 across in-domain examples (as seen in \Cref{fig:gltr_train_arxiv_test_arxiv_davinci}).

\begin{figure*}[h!]
    \centering
    \includegraphics[width=16cm]{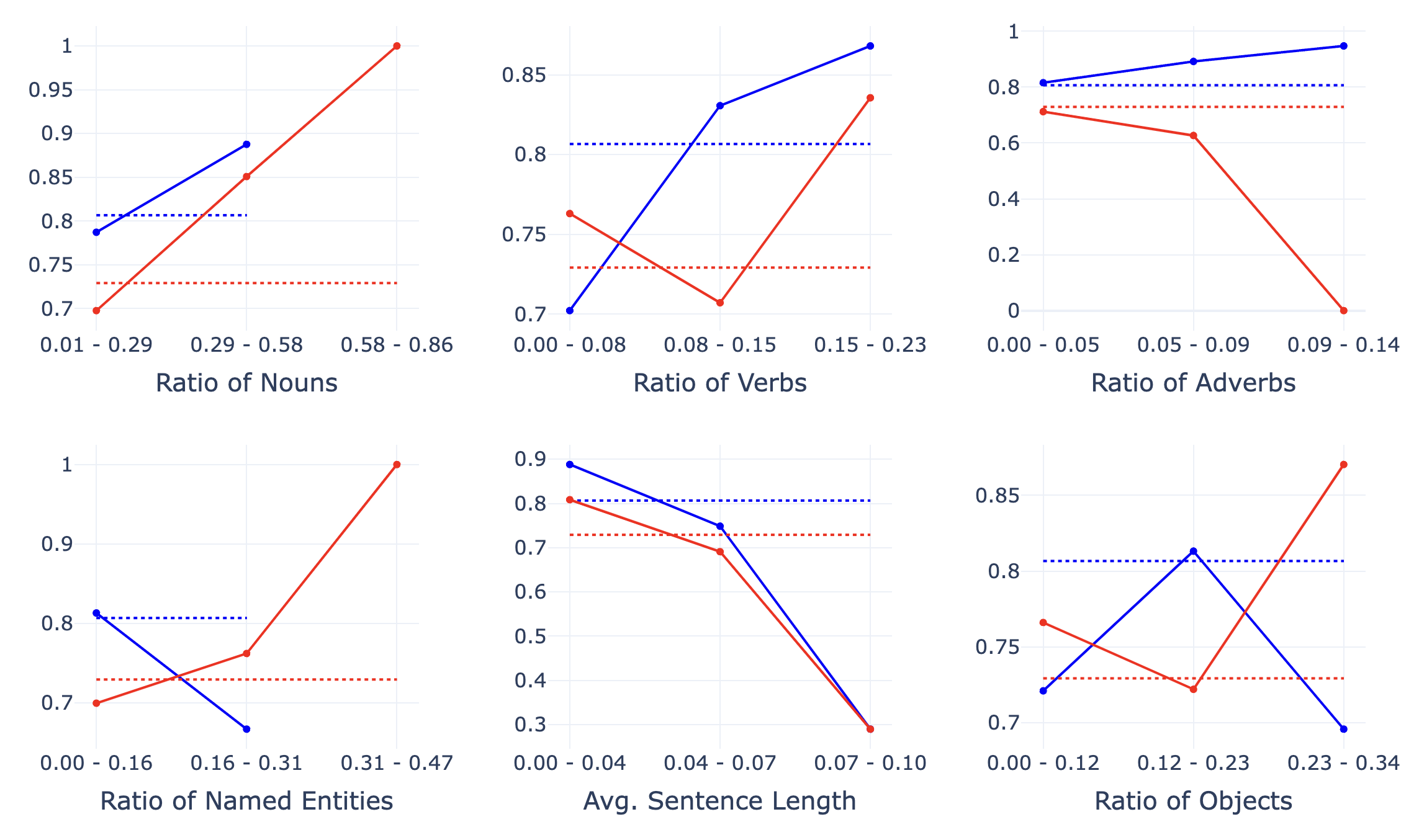}
    \caption{Binoculars Zero-Shot Detection. {\color{red}Red indicates machine-generated},
    {\color{blue}blue human-written} data, and dashed lines indicate baselines.}
    \label{fig:binoculars_zero_shot_detection}
\end{figure*}

\begin{figure*}[h!]
    \centering
    \includegraphics[width=16cm]{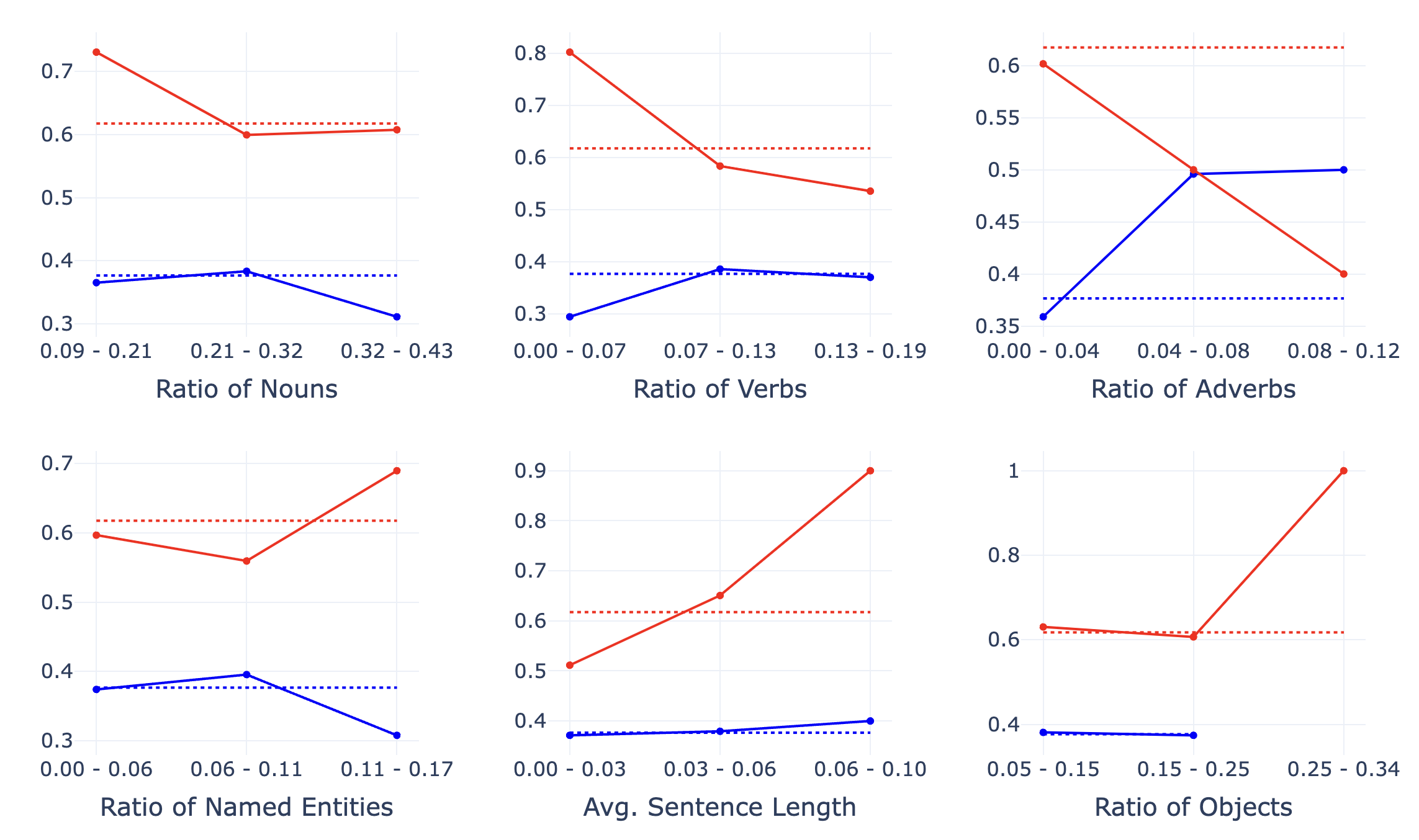}
    \caption{Llama 3.1 8b Zero-Shot Detection. {\color{red}Red indicates machine-generated},
    {\color{blue}blue human-written} data, and dashed lines indicate baselines.}
    \label{fig:llama_zero_shot_detection}
\end{figure*}

The results in \Cref{fig:binoculars_zero_shot_detection} and \Cref{fig:llama_zero_shot_detection} indicate that zero-shot detection methods for identifying machine-generated text, such as Binoculars, are heavily dependent on the length of the sentence. When evaluating longer sentences, the F scores degrade from around 0.9 to 0.3 across both classes. In certain adverb ratios, the F-score drops to 0 for machine-generated text. This suggests that zero-shot detection methods fixate on what are believed to be common features of machine-generated text (longer sentences and more adverbs).

% ==========================
% ===========================

% ==========================
% ===========================

% ==========================
% ===========================

% ==========================
% ===========================

\end{document}